\def\year{2016}
\documentclass[letterpaper]{article}
\usepackage{aaai}
\usepackage{times}
\usepackage{helvet}
\usepackage{courier}

\usepackage{url}
\usepackage{latexsym}
\usepackage{graphicx}
\newcommand{\argmax}{\arg\!\max}
\usepackage{amsmath}
\usepackage{amssymb}
\usepackage{subfigure}
\usepackage{multirow}
\usepackage{soul}

\begin{document}
%
\title{Learning to Answer Questions From Image Using Convolutional Neural Network}
\author{Lin~Ma~~~~~~~~Zhengdong~Lu~~~~~~~~Hang~Li\\
Noah's Ark Lab, Huawei Technologies\\
\texttt{forest.linma@gmail.com}~~~~~~~~\texttt{Lu.Zhengdong@huawei.com}~~~~~~~~\texttt{HangLi,HL@huawei.com}\\
}
\maketitle

\begin{abstract}

In this paper, we propose to employ the convolutional neural network (CNN) for the image question answering (QA). Our proposed CNN provides an end-to-end framework with convolutional architectures for learning not only the image and question representations, but also their inter-modal interactions to produce the answer. More specifically, our model consists of three CNNs: one image CNN to encode the image content, one sentence CNN to compose the words of the question, and one multimodal convolution layer to learn their joint representation for the classification in the space of candidate answer words. We demonstrate the efficacy of our proposed model on the DAQUAR and COCO-QA datasets, which are two benchmark datasets for the image QA, with the performances significantly outperforming the state-of-the-art.
\end{abstract}

\section{Introduction}
\label{sect:introduction}

Recently, the multimodal learning between image and language \cite{Ma:iccv2015,Makamura:IROS2013,XuXiChAAAI2015} has become an increasingly popular research area of artificial intelligence (AI). In particular, there have been rapid progresses on the tasks of bidirectional image and sentence retrieval \cite{frome:13,socher:14,klein:15,Karpathy:14a,Ordonez:11}, and automatic image captioning \cite{Chen:14,Karpathy:14b,Donahue:14,Fang:14,Kiros:14a,Krios:14b,klein:15,Mao:14b,Mao:14a,Vinyals:14,Xu:15}. In order to further advance the multimodal learning and push the boundary of AI research, a new ``AI-complete'' task, namely the visual question answering (VQA) \cite{Antol:2015} or image question answering (QA) \cite{Malinowski:2014a,Malinowski:2014b,Malinowski:2015a,Malinowski:2015b,Ren:2015},  is recently proposed. Generally, it takes an image and a free-form, natural-language like question about the image as the input and produces an answer to the image and question.

\begin{figure}
  \centering
  \includegraphics[width=\columnwidth]{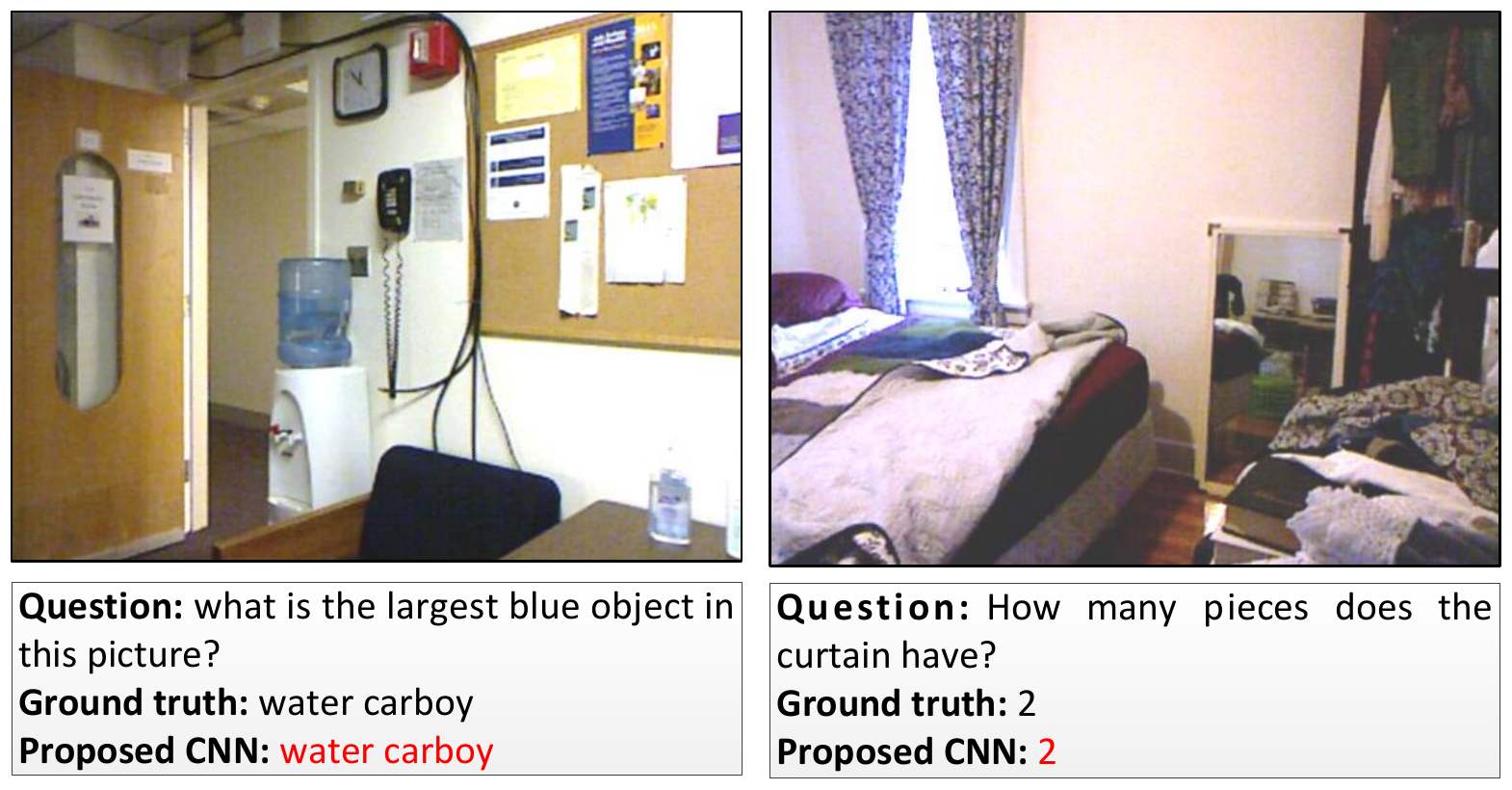}\\
  \caption{Samples of the image, the related question, and the ground truth answer, as well as the answer produced by our proposed CNN model.}
  \vspace{-10pt}
  \label{fig:imageqa_demo}
\end{figure}

Image QA differs with the other multimodal learning tasks between image and sentence, such as the automatic image captioning. The answer produced by the image QA needs to be conditioned on both the image and question. As such, the image QA involves more interactions between image and language. As illustrated in Figure \ref{fig:imageqa_demo}, the image contents are complicated, containing multiple different objects. The questions about the images are very specific, which requires a detailed understanding of the image content. For the question ``\texttt{\small what is the largest blue object in this picture?}'', we need not only identify the blue objects in the image but also compare their sizes to generate the correct answer. For the question ``\texttt{\small how many pieces does the curtain have?}'', we need to identify the object  ``\texttt{\small curtain}'' in the non-salient region of the image and figure out its quantity.

A successful image QA model needs to be built upon good representations of the image and question. Recently, deep neural networks have been used to learn image and sentence representations. In particular, convolutional neural networks (CNNs) are extensively used to learn the image representation for image recognition \cite{Simonyan:14,Szegedy:2014}. CNNs \cite{Hu:2014,Kim:2014,Kalchbrenner:2014}
also demonstrate their powerful abilities on the sentence representation for paraphrase, sentiment analysis, and so on. Moreover, deep neural networks \cite{Mao:14b,Karpathy:14a,Karpathy:14b,Vinyals:14} are used to capture the relations between image and sentence for image captioning and retrieval. However, for the image QA task, the ability of CNN has not been studied.

In this paper, we employ CNN to address the image QA problem. Our proposed CNN model, trained on a set of triplets consisting of (image, question, answer),  can answer free-form, natural-language like questions about the image. Our main contributions are:
\begin{enumerate}
  \item We propose an end-to-end CNN model for learning to answer questions about the image. Experimental results on public image QA datasets show that our proposed CNN model surpasses the state-of-the-art.
  \item We employ convolutional architectures to encode the image content, represent the question, and learn the interactions between the image and question representations, which are jointly learned to produce the answer conditioning on the image and question.
\end{enumerate}

\section{Related Work}
\label{sect:related_work}


Recently, the visual Turing test, an open domain task of question answering based on real-world images, has been proposed to resemble the famous Turing test. In \cite{Gao:2015} a human judge will be presented with an image, a question, and the answer to the question by the computational models or human annotators. Based on the answer, the human judge needs to determine whether the answer is given by a human (i.e. pass the test) or a machine (i.e. fail the test). Geman et al. \cite{Geman:2015} proposed to produce a stochastic sequence of binary questions from a given test image, where the answer to the question is limited to yes/no. Malinowski et al. \cite{Malinowski:2014b,Malinowski:2015a} further discussed the associated challenges and issues with regard to visual Turing test, such as the vision and language representations, the common sense knowledge, as well as the evaluation.

The image QA task, resembling the visual Turing test, is then proposed. Malinowski et al. \cite{Malinowski:2014a} proposed a multi-world approach that conducts the semantic parsing of question and segmentation of image to produce the answer. Deep neural networks are also employed for the image QA task, which is more related to our research work. The work  by \cite{Malinowski:2015b,Gao:2015} formulates the image QA task as a generation problem. Malinowski et al.'s model \cite{Malinowski:2015b}, namely the Neural-Image-QA, feeds the image representation from CNN and the question into the long-short term memory (LSTM) to produce the answer. This model ignores the different characteristics of questions and answers. Compared with the questions, the answers tend to be short, such as one single word denoting the object category, color, number, and so on. The deep neural network in \cite{Gao:2015}, inspired by the multimodal recurrent neural networks model \cite{Mao:14a,Mao:14b}, used two LSTMs for the representations of question and answer, respectively. In \cite{Ren:2015}, the image QA task is formulated as a classification problem, and the so-called visual semantic embedding (VSE) model is proposed. LSTM is employed to jointly model the image and question by treating the image as an independent word, and appending it to the question at the beginning or ending position. As such, the joint representation of image and question is learned, which is further used for classification. However, simply treating the image as an individual word cannot help effectively exploit the complicated relations between the image and question. Thus, the accuracy of the answer prediction may not be ensured. In order to cope with these drawbacks, we proposed to employ an end-to-end convolutional architectures for the image QA to capture the complicated inter-modal relationships as well as the representations of image and question. Experimental results demonstrate that the convolutional architectures can  can achieve better performance for the image QA task.

\section{Proposed CNN for Image QA}

\begin{figure}
  \centering
  \includegraphics[width=\columnwidth]{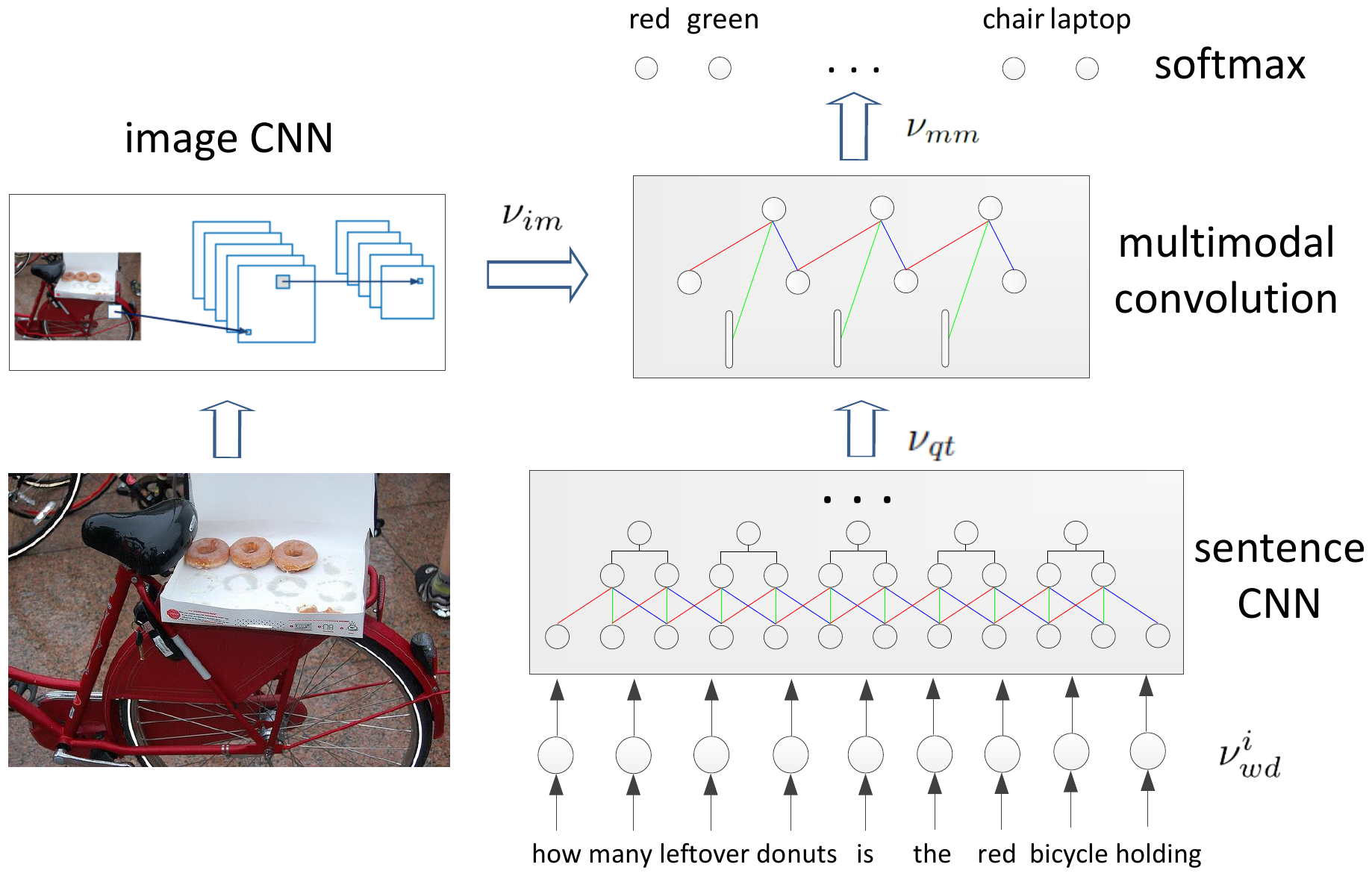}\\
  \caption{The proposed CNN model for image QA.}
  \label{fig:imageqa}
\end{figure}

For image QA, the problem is to predict the answer $a$ given the question $q$ and the related image $I$:
\begin{equation}\label{eq:formulation}
  a = \argmax_{a\in \Omega}~p(a|q,I;{\theta}),
\end{equation}
where $\Omega$ is the set containing all the answers. ${\theta}$ denotes all the parameters for performing image QA. In order to make a reliable prediction of the answer, the question $q$ and image $I$  need to be adequately represented.  Based on their representations, the relations between the two multimodal inputs are further learned to produce the answer.  In this paper, the ability of CNN is exploited for not only modeling image and sentence individually, but also capturing the relations and interactions between them.

As illustrated in Figure \ref{fig:imageqa}, our proposed CNN framwork for image QA consists of three individual CNNs: one image CNN encoding the image content, one sentence CNN generating the question representation, one multimodal convolution layer fusing the image and question representations together and generate the joint representation. Finally, the joint representation is fed into a softmax layer to produce the answer. The three CNNs and softmax layer are fully coupled for our proposed end-to-end image QA framework, with all the parameters (three CNNs and softmax) jointly learned in an end-to-end fashion.

\subsection{Image CNN}
\label{sect:image_CNN}
There are many research papers employing CNNs to generate image representations, which achieve the state-of-the-art performances on image recognition  \cite{Simonyan:14,Szegedy:2014}. In this paper, we employ the work \cite{Simonyan:14} to encode the image content for our image QA model:
\begin{equation}
\label{eq:imagecnn}
  \nu_{im} = \sigma (\mathbf{w}_{im}(CNN_{im}(I))+b_{im}),
\end{equation}
where $\sigma$ is a nonlinear activation function, such as Sigmoid and ReLU \cite{Dahl:2013}. $CNN_{im}$ takes the image as the input and outputs a fixed length vector as the image representation. In this paper, by chopping out the top softmax layer and the last ReLU layer of the CNN \cite{Simonyan:14}, the output of the last fully-connected layer is deemed as the image representation, which is a fixed length vector with dimension as 4096. Note that $\mathbf{w}_{im}$ is a mapping matrix of the dimension $d\times4096$, with $d$ much smaller than $4096$. On one hand, the dimension of the image representation is reduced from 4096 to $d$. As such, the total number of parameters for further fusing image and question, specifically the multimodal convolution process, is significantly reduced. Consequently, fewer samples are needed for adequately training our CNN model. On the other hand, the image representation is projected to a new space, with the nonlinear activation function $\sigma$ increasing the nonlinear modeling property of the image CNN. Thus its capability for learning complicated representations is enhanced. As a result, the multimodal convolution layer (introduced in the following section) can better fuse the question and image representations together and further exploit their complicated relations and interactions to produce the answer.

\subsection{Sentence CNN}
\label{sect:sentence_CNN}

\begin{figure}
  \centering
  \includegraphics[width=\columnwidth]{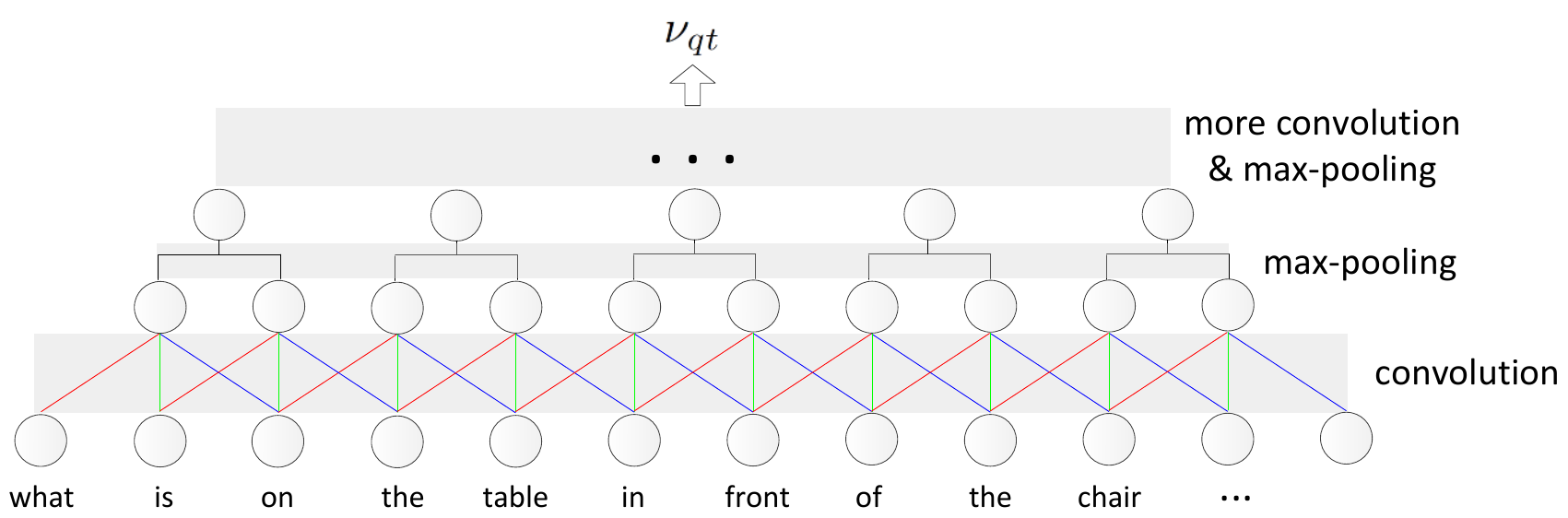}\\
  \caption{The sentence CNN for the question representation.}
  \label{fig:sentence_CNN}
\end{figure}

In this paper, CNN is employed to model the question for image QA. As most convolution models \cite{LeCun:95,Kalchbrenner:2014}, we consider the convolution unit with a local ``receptive field'' and shared weights to capture the rich structures and composition properties between consecutive words. The sentence CNN for generating the question representation is illustrated in Figure \ref{fig:sentence_CNN}. For a given question with each word represented as the word embedding \cite{Mikolov:13}, the sentence CNN with several layers of convolution and max-pooling is performed to generate the question representation $\nu_{qt}$.

\vspace{-10pt}
\paragraph{Convolution} For a sequential input $\nu$, the convolution unit for feature map of type-$f$  on the $\ell^{th}$ layer is
\begin{equation}
\label{eq_convolution}
\nu_{(\ell, f)}^{i} \overset{\text{def}}{=}
\sigma(\mathbf{w}_{(\ell,f)} \vec{\nu}_{(\ell-1)}^{i} + b_{(\ell,f)}),
\end{equation}
where $\mathbf{w}_{(\ell, f)}$ are the parameters for the $f$ feature map on the $\ell^{th}$ layer, $\sigma$ is the nonlinear activation function, and $\vec{\nu}_{(\ell-1)}^{i}$ denotes the segment of $(\ell\hspace{-1pt}-\hspace{-1pt}1)^{th}$ layer for the convolution at location $i$ , which is defined as follows.
\begin{equation}
\label{eq_segment}
\vec{\nu}_{(\ell-1)}^{i}  \overset{\text{def}}{=}  \nu_{(\ell-1)}^{i} \parallel \nu_{(\ell-1)}^{i+1} \parallel \cdots \parallel\nu_{(\ell-1)}^{i+s_{rp}-1},
\end{equation}
where $s_{rp}$ defines the size of local ``receptive field'' for convolution. ``$\parallel$'' concatenates the  $s_{rp}$  vectors into a long vector. In this paper, $s_{rp}$ is chosen as 3 for the convolution process. The parameters within the convolution unit are shared for the whole question with a window covering 3 semantic components sliding from the beginning to the end. The input of  the first convolution layer for the sentence CNN is the word embeddings of the question:
\begin{equation}
\label{eq_sentence_input}
\vec{\nu}_{(0)}^{i}  \overset{\text{def}}{=}  \nu_{wd}^{i} \parallel \nu_{wd}^{i+1} \parallel  \cdots \parallel\nu_{wd}^{i+s_{rp}-1},
\end{equation}
where $\nu_{wd}^{i}$ is the word embedding of the $i^{th}$ word in the question.

\vspace{-10pt}
\paragraph{Max-pooling} With the convolution process, the sequential $s_{rp}$ semantic components are composed to a higher semantic representation. However, these compositions may not be the meaningful representations, such as ``\texttt{\small is on the}'' of the question in Figure \ref{fig:sentence_CNN}. The max-pooling process following each convolution process is performed:
\begin{equation}
\label{eq_maxpooling}
\begin{split}
\nu^i_{(\ell+1,f)} = \max(\nu^{2i}_{(\ell,f)},\nu^{2i+1}_{(\ell,f)}).
\end{split}
\end{equation}
Firstly, together with the stride as two, the max-pooling process shrinks half of the representation, which can quickly make the sentence representation. Most importantly, the max-pooling process can select the meaningful compositions while filter out the unreliable ones. As such, the meaningful composition ``\texttt{\small of the chair}'' is more likely to be pooled out, compared with the composition ``\texttt{\small front of the}''.

The convolution and max-pooling processes exploit and summarize the local relation signals between consecutive words. More layers of convolution and max-pooling can help to summarize the local interactions between words at larger scales and finally reach the whole representation of the question. In this paper, we employ three layers of convolution and max-pooling to generate the question representation $\nu_{qt}$.

\subsection{Multimodal Convolution Layer}
\label{sect:multimodal_CNN}

\begin{figure}
 \centering
  \centering
  \includegraphics[width=0.75\columnwidth]{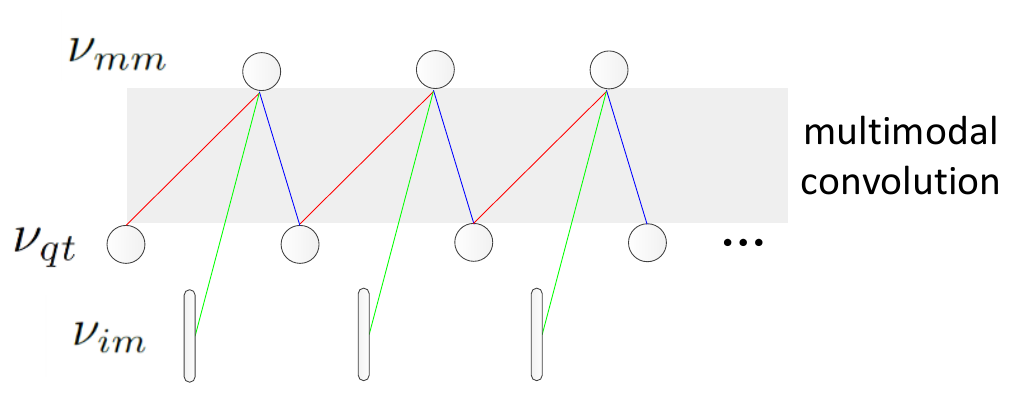}\\
  \caption{The multimodal convolution layer to fuse the image and question representations.}
  \label{fig:multimodal_CNN}
\end{figure}

The image representation $\nu_{im}$ and question representation $\nu_{qt}$ are obtained by the image and sentence CNNs, respectively. We design a new multimodal convolution layer on top of them, as shown in Figure \ref{fig:multimodal_CNN}, which fuses the multimodal inputs together to generate their joint representation for further answer prediction. The image representation is treated as an individual semantic component. Based on the image representation and the two consecutive semantic components from the question side, the mulitmodal convolution is performed, which is expected to capture the interactions and relations between the two multimodal inputs.
\begin{equation}
\label{eq_mmcnn_input}
\vec{\nu}_{mm}^{in}  \overset{\text{def}}{=}  \nu_{qt}^{i} \parallel \nu_{im} \parallel \nu_{qt}^{i+1},
\end{equation}
\vspace{-10pt}
\begin{equation}
\label{eq_mmcnn}
\nu_{(mm, f)}^{i} \overset{\text{def}}{=}
\sigma(\mathbf{w}_{(mm,f)} \vec{\nu}_{mm}^{in} + b_{(mm,f)}),
\end{equation}
where $\vec{\nu}_{mm}^{in}$ is the input of the multimodal convolution unit. $\nu_{qt}^{i}$ is the segment of the question representation at location $i$. $\mathbf{w}_{(mm,f)}$ and $b_{(mm,f)}$ are the parameters for the type-$f$ feature map of the multimodal convolution layer.

Alternatively, LSTM could be used to fuse the image and question representations, as in \cite{Malinowski:2015b,Ren:2015}.  For example, in the latter work, a bidirectional LSTM \cite{Ren:2015} is employed by appending the image representation to the beginning or ending position of the question. We argue that it is better to employ CNN than LSTM for the image QA task, due to the following reason, which has also been verified in the following experiment section. The relations between image and question are complicated. The image may interact with the high-level semantic representations composed from of a number of words, such as  ``\texttt{\small the red bicycle}''  in Figure \ref{fig:imageqa}. However, LSTM cannot effectively capture such interactions. Treating the image representation as an individual word, the effect of image will vanish at each time step of LSTM in \cite{Ren:2015}. As a result, the relations between the image and the high-level semantic representations of words may not be well exploited. In contrast, our CNN model can effectively deal with the problem. The sentence CNN first compose the question into a high-level semantic representations. The multimodal convolution process further fuse the semantic representations of image and question together and adequately exploit their interactions.

After the mutlimodal convolution layer, the multimodal representation $\nu_{mm}$ jointly modeling the image and question is obtained. $\nu_{mm}$ is then fed into a softmax layer as shown in Figure \ref{fig:imageqa}, which produces the answer to the given image and question pair.

\section{Experiments}
In this section, we firstly introduce the configurations of our CNN model for image QA and how we train the proposed CNN model. Afterwards, the public image QA datasets and evaluation measurements are introduced. Finally, the experimental results are presented and analyzed.

\subsection{Configurations and Training}
Three layers of convolution and max-pooling are employed for the sentence CNN. The numbers of the  feature maps for the three convolution layers are 300, 400, and 400, respectively. The sentence CNN is designed on a fixed architecture, which needs to be set to accommodate the maximum length of the questions. In this paper, the maximum length of the question is chosen as 38. The word embeddings are obtained by the skip-gram model \cite{Mikolov:13} with the dimension as 50. We use the VGG \cite{Simonyan:14} network as the image CNN. 
The dimension of $\nu_{im}$ is set as 400. The multimodal CNN takes the image and sentence representations as the input and generate the joint representation with the number of feature maps as 400.

The proposed CNN model is trained with stochastic gradient descent with mini batches of 100 for optimization, where the negative log likelihood is chosen as the loss. During the training process, all the parameters are tuned, including the parameters of nonlinear image mapping, image CNN, sentence CNN, multimodal convolution layer, and  softmax layer. Moreover, the word embeddings are also fine-tuned. In order to prevent overfitting, dropout (with probability 0.1) is used.

\subsection{Image QA Datasets}

We test and compare our proposed CNN model on the public image QA databases, specifically the DAQUAR \cite{Malinowski:2014a} and COCO-QA \cite{Ren:2015} datasets.

\noindent \textbf{DAQUAR-All} \cite{Malinowski:2014a}  This dataset consists of 6,795 training and 5,673 testing samples, which are generated from 795 and 654 images, respectively. The images are from all the 894 object categories. There are mainly three types of questions in this dataset, specifically the object type, object color, and number of objects. The answer may be a single word or multiple words.

\noindent \textbf{DAQUAR-Reduced} \cite{Malinowski:2014a} This dataset is a reduced version of DAQUAR-All, comprising 3,876 training and 297 testing samples. The images are constrained to 37 object categories. Only 25 images are used for the testing sample generation. Same as the DAQUAR-All dataset, the answer may be a single word or multiple words.

\noindent \textbf{COCO-QA} \cite{Ren:2015} This dataset consists of 79,100 training and 39,171 testing samples, which are generated from about 8,000 and 4,000 images, respectively. There are four types of questions, specifically the object, number, color, and location. The answers are all single-word.

\subsection{Evaluation Measurements}

One straightforward way for evaluating image QA is to utilize accuracy, which measures the proportion of the correctly answered testing questions to the total testing questions. Besides accuracy, Wu-Palmer similarity (WUPS) \cite{Wu:1994,Malinowski:2014a} is also used to measure the performances of different models on the image QA task. WUPS calculates the similarity between two words based on their common subsequence in a taxonomy tree. A threshold parameter is required for the calculation of WUPS. Same as the previous work \cite{Ren:2015,Malinowski:2014a,Malinowski:2015b}, the threshold parameters 0.0 and 0.9 are used for the measurements WUPS@0.0 and WUPS@0.9, respectively.

\subsection{Experimental Results and Analysis}

\subsubsection{Competitor Models}
We compare our models with recently developed models for the image QA task, specifically the multi-world approach \cite{Malinowski:2014a}, the VSE model \cite{Ren:2015}, and the Neural-Image-QA approach \cite{Malinowski:2015b}.

\subsubsection{Performances on Image QA}

\begin{table} \scriptsize
\vspace{-5pt}
\caption{Image QA performances on DAQUAR-All.}
\label{table:DAQUAR_all}
\begin{center}

    \renewcommand{\multirowsetup}{\centering}
    \begin{tabular}{|lccc|}
    \hline

      & \multirow{2}{*}{Accuracy} & WUPS & WUPS \\
      &                           & @0.9 & @0.0 \\
        \hline
        \hline
Multi-World Approach & \multirow{2}{*}{7.86} & \multirow{2}{*}{11.86} & \multirow{2}{*}{38.79} \\
\cite{Malinowski:2014a} & & &\\
\hline
\hline

Human Answers& \multirow{2}{*}{50.20} & \multirow{2}{*}{50.82} & \multirow{2}{*}{67.27} \\
\cite{Malinowski:2015b} & & &\\
Human Answers without image & \multirow{2}{*}{11.99} & \multirow{2}{*}{16.82} & \multirow{2}{*}{33.57} \\
\cite{Malinowski:2015b} & & &\\
\hline
\hline

Neural-Image-QA& & & \\
\cite{Malinowski:2015b} &&& \\
-multiple words&{17.49} & {23.28} & {57.76}\\
-single word  & 19.43 & 25.28 & 62.00  \\

Language Approach &&&\\
-multiple words&{17.06} & {22.30} & {56.53}\\
-single word  & 17.15& 22.80 & 58.42  \\
\hline
\hline

Proposed CNN &&& \\
-multiple words & \textbf{21.47} & \textbf{27.15} & \textbf{59.44} \\
-single word & \textbf{24.49} & \textbf{30.47} & \textbf{66.08} \\
\hline

    \end{tabular}

\end{center}
\vspace{-20pt}

\end{table}

The performances of our proposed CNN model on the DAQUAR-All, DAQUAR-Reduced, and COCO-QA datasets are illustrated in Table \ref{table:DAQUAR_all}, \ref{table:DAQUAR_Reduced}, and \ref{table:COCO_QA}, respectively. For DAQUAR-All and DAQUAR-Reduced datasets with multiple words as the answer to the question, we treat the answer comprising multiple words as an individual class for training and testing.

\begin{table} \scriptsize
\caption{Image QA performances on DAQUAR-Reduced.}
\label{table:DAQUAR_Reduced}
\begin{center}

    \renewcommand{\multirowsetup}{\centering}
    \begin{tabular}{|lccc|}
    \hline

      & \multirow{2}{*}{Accuracy} & WUPS & WUPS \\
      &                           & @0.9 & @0.0 \\
        \hline
        \hline
Multi-World Approach & \multirow{2}{*}{12.73} & \multirow{2}{*}{18.10} & \multirow{2}{*}{51.47} \\
\cite{Malinowski:2014a} & & &\\
\hline
\hline

Neural-Image-QA & & & \\
\cite{Malinowski:2015b} &&& \\
-multiple words&{29.27} & {36.50} & {79.47}\\
-single word  & 34.68& 40.76 & 79.54  \\

Language Approach &&&\\
-multiple words&{32.32} & {38.39} & {80.05}\\
-single word  & 31.65& 38.35 & 80.08  \\
\hline
\hline

VSE \cite{Ren:2015}  &&&\\
-single word & &  &  \\
GUESS & 18.24 & 29.65 & 77.59 \\
BOW & 32.67 & 43.19 & 81.30 \\
LSTM & 32.73 & 43.50 & 81.62 \\
IMG+BOW & 34.17 & 44.99 & 81.48\\
VIS+LSTM & 34.41 & 46.05 & 82.23 \\
2-VIS+BLSTM & 35.78 & 46.83 & 82.14 \\
\hline
\hline

Proposed CNN &&& \\
-multiple words & \textbf{38.38} & \textbf{43.43} & \textbf{80.63} \\
-single word & \textbf{42.76} & \textbf{47.58} & \textbf{82.60} \\
\hline

    \end{tabular}

\end{center}

\vspace{-20pt}
\end{table}

For the DAQUAR-All dataset, we evaluate the performances of different image QA models on the full set (``multiple words''). The answer to the image and question pair may be a single word or multiple words.  Same as the work \cite{Malinowski:2015b}, a subset containing the samples with only a single word as the answer is created and employed for comparison (``single word''). Our proposed CNN model significantly outperforms the multi-world approach and Neural-Image-QA in terms of accuracy, WUPS@0.0, and WUPS@0.9. Specifically, our proposed CNN model achieves over $20\%$ improvement compared to Neural-Image-QA in terms of accuracy on both ``multiple words'' and ``single word''.  
The results, shown in Table. \ref{table:DAQUAR_all}, demonstrate that our CNN model can more accurately model the image and question as well as their interactions, thus yields better performances for the image QA task.
Moreover, the language approach \cite{Malinowski:2015b}, which only resorts to the question performs inferiorly to the approaches that jointly model the image and question. The image component is thus of great help to the image QA task. One can also see that the performances on ``multiple words'' are generally inferior to those on ``single word''.


For the DAQUAR-Reduced dataset, besides the Neural-Image-QA approach, the VSE model is also compared on ``single word''. Moreover, some of the methods introduced in \cite{Ren:2015} are also reported and compared. GUESS is the model which randomly outputs the answer according to the question type. BOW treats each word of the question equally and sums all the word vectors to predict the answer by logistic regression. LSTM is performed only on the question without considering the image, which is similar to the language approach \cite{Malinowski:2015b}. IMG+BOW performs the multinomial logistic regression based on the image feature and a BOW vector obtained by summing all the word vectors of the question. VIS+LSTM and 2-VIS+BLSTM  are two versions of the VSE model. VIS+LSTM has only a single LSTM to encode the image and question in one direction, while 2-VIS+BLSTM uses a bidirectional LSTM to encode the image and question along with both directions to fully exploit the interactions between image and each word of the question. It can be observed that 2-VIS+BLSTM outperforms VIS+LSTM with a big margin. The same observation can also be found on the COCO-QA dataset, as shown in Table \ref{table:COCO_QA}, demonstrating that the bidirectional LSTM can more accurately model the interactions between image and question than the single LSTM. Our proposed CNN model significantly outperforms the competitor models. More specifically, for the case of ``single word'', our proposed CNN achieves nearly $20\%$ improvement in terms of accuracy over the best competitor model 2-VIS+BLSTM.


\begin{table} \scriptsize
\caption{Image QA performances on COCO-QA.}
\label{table:COCO_QA}
\begin{center}

    \renewcommand{\multirowsetup}{\centering}
    \begin{tabular}{|lccc|}
    \hline

      & \multirow{2}{*}{Accuracy} & WUPS & WUPS \\
      &                           & @0.9 & @0.0 \\
        \hline
        \hline

VSE \cite{Ren:2015}  &&&\\
GUESS & 6.65 & 17.42 & 73.44 \\
BOW & 37.52 & 48.54 & 82.78 \\
LSTM & 36.76 & 47.58 & 82.34 \\
IMG & 43.02 & 58.64 & 85.85 \\
IMG+BOW & 55.92 & 66.78 & 88.99 \\
VIS+LSTM & 53.31 & 63.91 & 88.25 \\
2-VIS+BLSTM & 55.09 & 65.34 & 88.64 \\
FULL & 57.84 & 67.90 & 89.52 \\
\hline
\hline
Proposed CNN without & \multirow{2}{*}{56.77} & \multirow{2}{*}{66.76} & \multirow{2}{*}{88.94} \\
multimodal convolution layer & & &\\
Proposed CNN without & \multirow{2}{*}{37.84} & \multirow{2}{*}{48.70} & \multirow{2}{*}{82.92} \\
image representation & & &\\
Proposed CNN & \textbf{58.40} &  \textbf{68.50} & \textbf{89.67}  \\
\hline

    \end{tabular}
\end{center}
\vspace{-20pt}
\end{table}

For the COCO-QA dataset, IMG+BOW outperforms VIS+LSTM and 2-VIS+BLSTM, demonstrating that the simple multinomial logistic regression of IMG+BOW can better model the interactions between image and question, compared with the LSTMs of VIS+LSTM and 2-VIS+BLSTM. By averaging VIS+LSTM, 2-VIS+BLSTM, and IMG+BOW, the FULL model is developed, which summarizes the interactions between image and question from different perspectives thus yields a much better performance. As shown in Table \ref{table:COCO_QA}, our proposed CNN model outperforms all the competitor models in terms of  all the three evaluation measurements, even the FULL model. The reason may be that the image representation is of highly semantic meaning, which should interact with the high semantic components of the question. Our CNN model firstly uses the convolutional architectures to compose the words to highly semantic representations. Afterwards, we let the image meet the composed highly semantic representations and use convolutional architectures to exploit their relations and interactions for the answer prediction. As such, Our CNN model can well model the relations between image and question, and thus obtain the best performances.


\subsubsection{Influence of Multimodal Convolution Layer}

The image and question needs to be considered together for the image QA. The multimodal convolution layer in our proposed CNN model not only fuses the image and question representations together but also learns the interactions and relations between the two multimodal inputs for further question prediction. The effect of the multimodal convolution layer is examined as follows. The image and question representations are simply concatenated together as the input of the softmax layer for the answer prediction. We train the network in the same manner as the proposed CNN model. The results are provided in Table \ref{table:COCO_QA}.  Firstly, it can be observed that without the multimodal convolution layer, the performance on the image QA has dropped. Comparing to the simple concatenation process fusing the image and question representations, our proposed multimodal convolution layer can well exploit the complicated relationships between image and question representations. Thus a better performance for the answer prediction is achieved. Secondly, the approach without multimodal convolution layer outperforms the IMG+BOW, VIS+LSTM and 2-VIS+BLSTM, in terms of accuracy. The better performance is mainly attributed to the composition ability of the sentence CNN. Even with the simple concatenation process, the image representation and composed question representation can be fuse together for a better image QA model.


\subsubsection{Influence of Image CNN and Effectiveness of Sentence CNN}
As can be observed in Table \ref{table:DAQUAR_all}, without the image content, the accuracy of human answering the question drops from $50\%$ to $12\%$. Therefore, the image content is critical to the image QA task.  Same as the work \cite{Malinowski:2015b,Ren:2015}, we only use the question representation obtained from the sentence CNN to predict the answer. The results are listed in Table \ref{table:COCO_QA}. Firstly, without the use of image representation, the performance of our proposed CNN significantly drops, which again demonstrates the importance of image component to the image QA. Secondly, the model only consisting of the sentence CNN performs better than LSTM and BOW for the image QA. It indicates that the sentence CNN is more effective to generate the question representation for image QA, compared with LSTM and BOW. Recall that the model without multimodal convolution layers outperforms IMG+BOW, VIS+LSTM, and 2-VIS+BLSTM, as explained above. 
By incorporating the image representation, the better modeling ability of our sentence CNN is demonstrated.


Moreover, we examine the language modeling ability of the sentence CNN as follows. The words of the test questions are randomly reshuffled. Then the reformulated questions are sent to the sentence CNN to check whether the sentence CNN can still generate reliable question representations and make accurate answer predictions. For randomly reshuffled questions, the results on COCO-QA dataset are 40.74, 53.06, and 80.41 for the accuracy, WUPS@0.9, and WUPS@0.0, respectively, which are significantly inferior to that of natural-language like questions. The result indicates that the sentence CNN possesses the ability of modeling natural questions. The sentence CNN uses the convolution process to compose and summarize the neighboring words. And the reliable ones with higher semantic meanings will be pooled and composed further to reach the final sentence representation. As such, the sentence CNN can compose the natural-language like questions to reliable high semantic representations.


\section{Conclusion}

In this paper, we proposed one CNN model to address the image QA problem. The proposed CNN model relies on convolutional architectures to generate the image representation, compose consecutive words to the question representation, and learn the interactions and relations between the image and question for the answer prediction. Experimental results on public image QA datasets demonstrate the superiority of our proposed model over the state-of-the-art methods.

\section{Acknowledgement}
The work is partially supported by China National 973 project 2014CB340301. The authors are grateful to Baotian Hu and Zhenguo Li for their insightful discussions and comments. 

{
\bibliographystyle{aaai}
\bibliography{aaai}
}
\end{document}